# CNN-Based Structural Damage Detection using Time-Series Sensor Data

*Ishan Pathak[1], Ishan Jha[2], Aditya Sadana[3], Basuraj Bhowmik[4,*]*

[1]Member of Technical Staff, Vymo, Bangalore, India
[2]Research Scholar, Indian Institute of Technology (BHU), Varanasi, India
[3]Software Engineer, SmartReach.io, Hyderabad, India
[4]Assistant Professor, Indian Institute of Technology (BHU), Varanasi, India

## Abstract

*Structural Health Monitoring (SHM) is vital for evaluating structural condition, aiming to detect damage through sensor data analysis. It aligns with predictive maintenance in modern industry, minimizing downtime and costs by addressing potential structural issues. Various machine learning techniques have been used to extract valuable information from vibration data, often relying on prior structural knowledge. This research introduces an innovative approach to structural damage detection, utilizing a new Convolutional Neural Network (CNN) algorithm. In order to extract deep spatial features from time series data, CNNs are taught to recognize long-term temporal connections. This methodology combines spatial and temporal features, enhancing discrimination capabilities when compared to methods solely reliant on deep spatial features. Time series data are divided into two categories using the proposed neural network: undamaged and damaged. To validate its efficacy, the method's accuracy was tested using a benchmark dataset derived from a three-floor structure at Los Alamos National Laboratory (LANL). The outcomes show that the new CNN algorithm is very accurate in spotting structural degradation in the examined structure.*

## 1. Introduction

With the advent of computers, the era of computing began, ushering in a transformative wave that extended its influence to the field of civil engineering. This digital revolution made intricate calculations, such as structural analysis, optimization, and the emerging discipline of Structural Health Monitoring (SHM), significantly more accessible and streamlined in nature [1-3]. SHM has emerged as a pivotal discipline within civil engineering and infrastructure management. Its primary purpose lies in evaluating the state of structures, providing essential insights into their current condition, and enabling early detection of structural damage. In an era where minimizing operational downtime, optimizing resource allocation, and enhancing safety are paramount concerns, SHM assumes a crucial role in ensuring the efficient operation and maintenance of critical infrastructure. At its core, SHM empowers the proactive identification of structural anomalies and defects, forming the basis for predictive maintenance strategies. Predictive maintenance, a cornerstone of modern industrial operations, seeks to pre-empt unplanned downtime, curtail operational expenses, and safeguard the long-term integrity of structures. By identifying potential failures before they escalate into critical issues, precise maintenance schedules can be established, resources can be allocated efficiently, and costly emergency repairs can be averted. Recent advances in high-performance computing and low-cost sensor technologies have ushered in a new era, making continuous and effective SHM more attainable. Notably, vibration-based damage identification has seen significant advancements, underpinned by extensive research efforts. Diverse techniques, algorithms, and methodologies have emerged to address the multifaceted challenges encountered across structures of varying complexities. Detailed reviews encompassing different facets of SHM can be found in publications such as those authored by [4-9]

However, despite the substantial progress in the field of SHM, two major obstacles have impeded its widespread application in civil infrastructure systems. Firstly, a deficiency is present in effective and dependable methodologies for processing extensive volumes of diverse response data signals. Secondly, cost-effective sensors have been in short supply. In light of recent technological breakthroughs, a growing array of sensors and its networks now may be employed, generating copious amounts of data responses. Whilst this data might not consistently provide adequate or pertinent information for traditional SHM approaches, machine learning (ML) techniques that are data-driven have arisen as a favourable solution for evaluating the overall condition of host structures. Noteworthy contributions to this field include the work of [10-12].

Data-driven damage detection methods have gained traction in addressing the problem as pattern recognition tasks, wherein neural networks (NN) play a pivotal role. Traditional neural networks, however, face challenges related to the need for extensive training datasets and high computation costs. To address these constraints, recent endeavours have replaced conventional feature extraction and classification tasks in damage detection challenges with deep learning tools, utilizing both processed and raw signals with no need for manually crafted features. Exemplifying this shift, [13,14] have employed deep learning techniques to augment damage detection, overcoming training dataset constraints. Leading the field of deep learning during the age of big data are Convolutional Neural Networks (CNNs), possessing the capacity to acquire knowledge from extensive data derived from numerous observables (sensors). While CNNs have found success in diverse domains for example electrocardiogram signal classification [15] and image analysis [16], their adoption in SHM remains relatively nascent [17,18]. Encouragingly, the literature presents accomplished implementation of CNNs in SHM, including their effectiveness in detecting damage in steel frames [19], concrete crack monitoring and pavement [20,21], and evaluating the overall condition of the system [22]. The growing focus to deep learning methods, particularly CNN-based models with hierarchical feature learning capabilities, underscores their significant potential in addressing SHM challenges.

Although the bulk of SHM response data is gathered in the domain of time, few works have proposed using modified data for distinguishing damaged from undamaged structures by converting data from the time domain to either the frequency or time-frequency domains [23,24]. In order to reduce discrepancies in the vibration properties of finite element models and actual structures in damage localisation using vibration in structures, [25] presented a nature-inspired approach. Similar to this, [26] created a workable model for high-rise structures based on strain sensing. Through the examination of vibration response data, [27] created an approach based on the Stretching Methodology for the prolonged tracking of massive structures. A nonlinear model-data fusion approach for estimating the state in non-linear structural assemblies displaying hysteresis was recently proposed by [28]. This algorithm was applied to the monitor seismicity of both experimental and real-world buildings equipped with instrumentation. [28-30] presented a blind identification technique employing Sparse Component Analysis within a time-frequency framework. Utilizing recorded data on acceleration from Yonghe Bridge sensors, this method was experimentally evaluated.

Furthermore, [23,24] utilized machine learning-driven methods for detecting structural damage through vibration analysis in extensive bridge structures, making use of time-frequency approaches. Modal parameter identification found enhanced utility in the work of [31-33], who harnessed wavelet and Hilbert transforms. With regard to SHM applied to large-scale structures [34], improved the Bayesian substructure identification approach for solving inverse challenges. [35] presented a multitask sparse Bayesian learning method, which serves the dual purpose of data reconstruction and estimation of structural stiffness. Collectively, these strategies include feature extraction methods that are used to determine if, where, and what kind of damage is present in structural systems. See [36,7] for in-depth analyses of a variety of techniques, including those founded on statistical and probabilistic concepts.

Recent advancements have led to the development of self-powered sensors capable of acquiring acceleration or strain response data in a condensed manner [37-39]. These sensors employ memory cells to accumulate and retain the time duration of the measured responses, defined based on preselected thresholds and discretized. The recorded response data is presented as a histogram of events in a compressed format, as opposed to a traditional time-history of responses. To harness the potential of this concise, discrete, and finite response data towards detection of damage, it is imperative to develop new and inventive methodologies. With the rapid technological advancements, there is an abundance of data, including vibration data from SHM, that can be harnessed for analysis. Modern algorithms like deep learning have shown remarkable efficiency in handling massive datasets without the need for extensive feature extraction [40,41]. CNNs, a type of artificial neural network, have found applications in diverse domains, including image classification, speech recognition, and damage detection [42-45]. In damage detection, CNNs have found its use in processing time series data encoded as images, enabling the localization and classification of damage levels [42].

This research represents a stride towards addressing these challenges by introducing an innovative approach to structural damage detection. The method harnesses the capabilities of a new CNN algorithm. CNNs, initially designed for image processing, have showcased remarkable effectiveness in capturing intricate spatial features. In our adaptation, we employ a state-of-the-art CNN algorithm tailored specifically for SHM. By leveraging its deep learning capabilities, we extract detailed spatial features from sensor data. It also excels at capturing long-term temporal dependencies within time series data. This integration not only enhances the understanding of structural behaviour from the model but also bolsters its ability to make informed assessments based on the evolving patterns within the data.

The fusion of spatial and temporal features within the proposed methodology represents a significant departure from conventional methods that are primarily reliant on spatial features. This innovation promises a substantial enhancement in accuracy when it comes to structural damage detection, as it takes into account the evolving nature of structural conditions over time. To evaluate its effectiveness, a series of tests is conducted using a benchmark dataset derived from a three-storey structure at the esteemed Los Alamos National Laboratory (LANL). The findings underscore the potential of the new CNN algorithm to significantly enhance structural health assessment by achieving a high degree of precision in identifying structural damage within the structure under examination.

The subsequent sections, delve into the intricate intricacies of the methodology, present the results of experimentation, and engage in a detailed discussion, providing a profound exploration of approach and its far-reaching ramifications for the field of Structural Health Monitoring.

## 2. Methodology

LeCun et al. [42] introduced CNN theory as a deep learning model. Convolution and pooling [44], two essential layers that may be coupled to fully-connected layers, make up the majority of CNN architecture. These layers generate feature maps as 2D matrices from the input data. One of CNN's major advantages is its capacity to learn relevant features from provided data while also benefiting from parameter sharing, resulting in significantly lower computational costs compared to other neural network classes. Despite the fact that CNN typically handles 2D matrix inputs, a modified framework known as 1D CNN was suggested for image processing. All the benefits of conventional CNN are still there when accommodating input data from a 1D matrix.

Due to its lower computational cost, 1D CNN has recently been shown to provide benefits when used with data from time series for SHM. In order to achieve computational efficiency, 1D CNN only uses array functions in forward as well as backward propagation. Additionally, shallow architecture It is simple and efficient to train 1D CNNs to acquire the skills required for time series applications. This study adapts [46] to use a 1D CNN architecture for damage identification. It comprises two primary layers: convolution and pooling, aimed at extracting relevant features from the input data. Fully connected layers are then used to classify the extracted features.

Forward propagation is determined by Eq. 1 for each layer:

$$l_k^l = f(\sum_{i=1}^{N_{l-1}} (x_i^{l-1}.w_{ik}^l) + b_k^l)$$ **Eq. (1)**

Where, l represents the layer, $x_i^{l-1}$ is the input from the previous layer, f(.) is the activation function, $w_{ik}^l$ is the kernel from the i$^{th}$ neuron in layer l-1 to the k$^{th}$ neuron in layer l, and $b_k^l$ is the bias of the k$^{th}$ neuron in layer. The intermediate output $l_k^l$ is calculated based on the activation function f(.).

By computing the slopes for the loss function E(y) in connection to the CNN weights, back-propagation is used to train the network. The formula in Eq. (2) is used to find the derivative of the error with regard to each weight:

$$\frac{dE}{dw_{ik}^l} = \Delta w_{ik}^l$$ **Eq. (2)**

The weight is then updated using the computed gradients of the layers, as depicted in Eq. (3):

$$w_{ik}^l = w_{ik}^l + \eta \Delta w_{ik}^l$$ **Eq. (3)**

Where, $w_{ik}^l$ represents the weight for the following iteration and $\eta$ represents the rate of learning. [45] provides further information about the weight update calculation.

Following is the description of a typical CNN:

1. Input Layer: The initial data, which is often an image, is represented by the input layer. For a colour image, this layer would have three channels (Red, Green, Blue - RGB), and the image's dimensions are specified. For example, a common input size for image classification might be 3x32x32 (3 channels, 32 pixels in width, and 32 pixels in height).
2. Convolutional Layers: In a CNN, convolutional layers are at its core. They consist of multiple filters (kernels) that convolve or slide over the input image to detect patterns and features. Each filter identifies different features by performing convolution operations. For example, in the first convolutional layer, you might have 32 filters, each with a size of 3x3 pixels. A feature map that depicts the presence of particular characteristics in the input is produced as a consequence of the convolution procedure.

3. Activation Function: An activation function, frequently a ReLU (Rectified Linear Unit), is executed element-wise after each convolution process to provide nonlinearity. This aids the network's modelling of intricate relationships.

4. Pooling Layers: The most crucial data is kept while the spatial dimensions for the feature maps are reduced using pooling layers, frequently known as MaxPooling or AveragePooling. Use of 2x2 pooling, for instance, is a typical arrangement where the largest (or mean) value across a 2x2 window for the feature map is chosen.

5. Fully Connected Layers: To create final predictions, fully connected layers are used after a number of convolutional and pooling layers. These layers are tightly coupled, which means that each neuron is linked to every other neuron in the layer above. These layers' neuronal densities might vary and are normally decided upon during design. For instance, the first completely linked layer can include 128 neurons.

6. Output Layer: The final prediction is generated by the output layer and is dependent on the particular job. The total amount of neurons in the output layer for image classification tasks equates to the amount of classes you are attempting to predict. For example, if you are classifying images into 10 categories, the output layer will have 10 neurons. Common activation functions in the output layer include Softmax for classification problems, which convert raw scores into class probabilities. Fig. 1 provides an illustration of a schematic depiction of CNN.

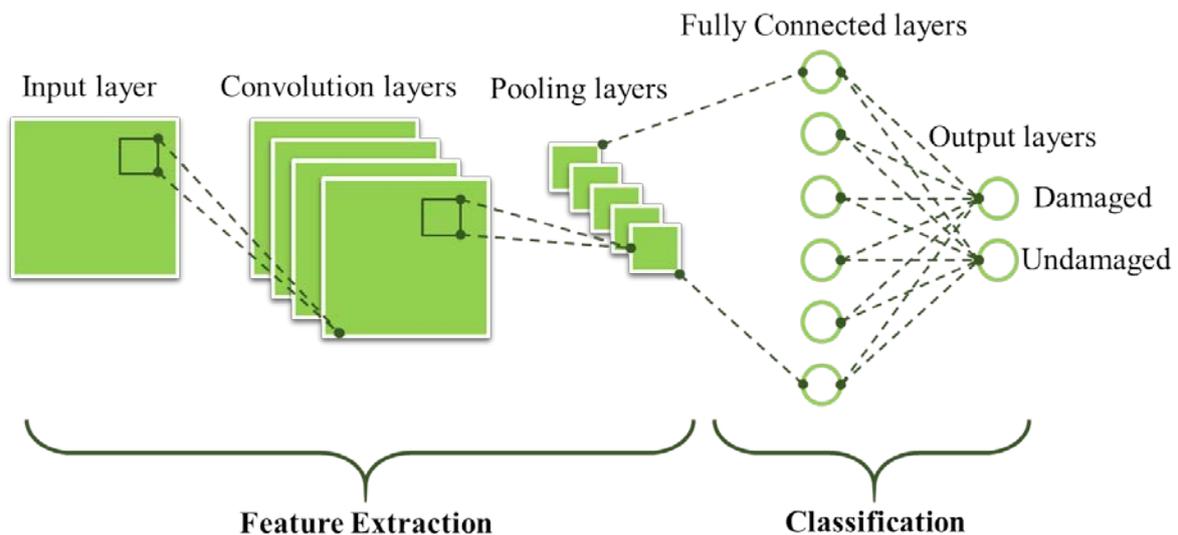

**Fig. 1:** Schematic depiction of CNN.

## 3. Experimental Description

In order to validate the method put forth in this study, a set of experimental data from the Los Alamos National Laboratory (LANL) is employed. These data pertain to a three-floor structural system.

Aluminium plates and columns are put together in the structural arrangement shown in Fig. 2 utilizing bolted connections, allowing the structure to move only in the x-direction along rails. Each floor of the system comprises four aluminium columns, measuring 17.7x2.5x0.6cm, connected to upper and lower aluminium plates measuring 30.5x30.5x2.5cm. This arrangement essentially constitutes a four-degree-of-freedom system. Furthermore, a central column of dimensions 15.0x2.5x2.5cm is suspended from the top floor. By interacting with a bumper positioned on the floor below, this centre column may produce nonlinear behaviour, enabling the simulation of damage. At particular excitation levels, the bumper's location may be changed to alter the severity of the impact. This source of damage emulates phenomena such as fatigue cracks undergoing cyclic opening and closing, or loose connections rattling when subjected to dynamic loads.

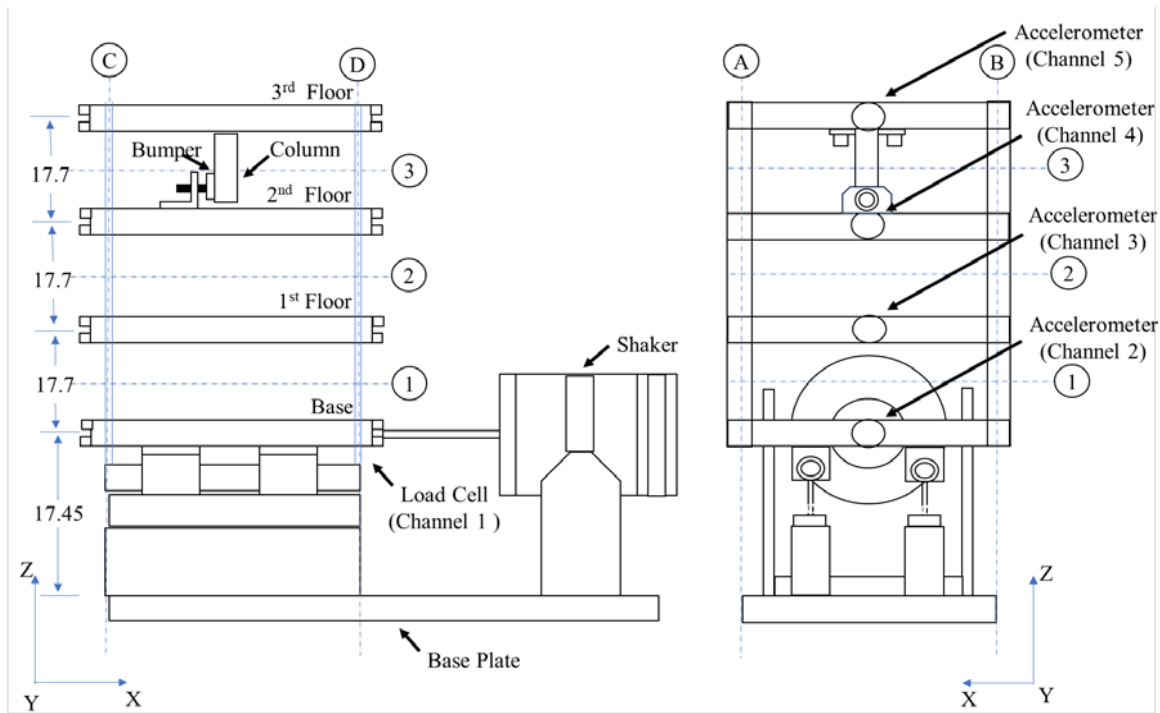

**Fig. 2:** Structural configuration of three-story building structure at LANL

## 3.1 Data Acquisition System

An electrodynamic shaker is used to apply a lateral stimulation through the base floor across the structure's centreline. Together, the framework and shaker are set onto an aluminium baseplate that is 76.2x30.5x2.5 cm in size, supported by rigid foam to minimize the introduction of unmeasured excitation from external sources. To measure the input force communicated from the shaker to the structure, a load cell (Channel 1) with a nominal sensitivity of 2.2 mV/N is attached to the end of a stinger. To measure the system's reaction, four accelerometers (Channels 2-5) with nominal sensitivity of 1000 mV/g are installed across the centreline of every floor, opposite from the excitation source. These accelerometers are resistant to the torsional modes of the structure since they are situated at the centerline of each level. To reduce torsional excitation, the shaker and linear bearings are positioned.

A PXI data collection system from National Instruments is used to assist data collecting and processing. A PXI-4461 DAQ module is used to create analog output waveforms, and a PXI-4472B DAQ module is used to record the response data of all five sensors. Employing a PCB 482A16 signal conditioner, the five sensor channels' ICP conditioning is accomplished. The Techron 5530 Power Supply Amplifier is linked to the analog output channel, which drives the shaker. The digitalized analog sensor signals are captured in blocks of 65536 points and digitize at a rate of 2560 Hz. The data is then downscaled to 8192 data points at intervals of 3.125 ms, which results in a sampling frequency of 320 Hz. These variables produce time series with a length of 25.6 seconds. The structure is stimulated using a band-limited random stimulation in the 20–150 Hz frequency range, which was selected to avoid activating the stiff body modes existent under 20 Hz. Within the National Instruments system, the excitation level is set at 2.6 V RMS. As shown in Table 1, the structural state conditions have been separated into four major groupings. It is crucial to remember that 50 experiments were run for each state condition, producing 50 time histories for each channel for every state condition.

**Table 1:** Structural state conditions.

| State#1 | Undamaged | Reference state with baseline condition |
|---|---|---|
| State#2 | Undamaged | Added 1.2 kg mass at the base |
| State#3 | Undamaged | Added 1.2 kg mass on the 1st floor |
| State#4 | Undamaged | Reduction of stiffness by 87.5% in column 1BD |
| State#5 | Undamaged | Reduction of stiffness by 87.5% in column 1AD and 1BD |
| State#6 | Undamaged | Reduction of stiffness by 87.5% in column 2BD |
| State#7 | Undamaged | Reduction of stiffness by 87.5% in column 2AD and 2BD |
| State#8 | Undamaged | Reduction of stiffness by 87.5% in column 3BD |
| State#9 | Undamaged | Reduction of stiffness by 87.5% in column 3AD and 3BD |
| State#10 | Damaged | 0.20mm Gap introduced |
| State#11 | Damaged | 0.15mm Gap introduced |
| State#12 | Damaged | 0.13mm Gap introduced |
| State#13 | Damaged | 0.10mm Gap introduced |
| State#14 | Damaged | 0.05mm Gap introduced |
| State#15 | Damaged | 0.20mm Gap introduced and added 1.2 kg mass at base |
| State#16 | Damaged | 0.20mm Gap introduced and added 1.2 kg mass on 1st floor |
| State#17 | Damaged | 0.10mm Gap introduced and added 1.2 kg mass on 1st floor |

The baseline situation is represented by the first group, State#1 in Table 1. Bumper along with hanging column are present in this scenario, however their distance from one another has been maintained to prevent any collisions during excitation. States with modelled operational and environmental unpredictability are included in the second group. Modifications to the mass distribution or stiffness of the structure are common causes of this fluctuation. Tests with different stiffness and mass-loading settings were carried out (State#2-9) to replicate these variations in operating and environmental situations. When the mass is located at the base, the first level and base required the addition of 1.2 kg (about 19% of the overall mass for each floor). By 87.5% lowering the stiffness of one or more columns, stiffness variations were added. To achieve this, a column with a cross-sectional width in the axis of shaking half that of the original column was used in its place.

The third category consists of damaged state circumstances, which are mimicked by adding nonlinearity to the system by varying the distance between the hanging column and the bumper. Varying the gap widths between them allowed for the introduction of various amounts of nonlinearity. The last set consists of state conditions that include simulated damage in along with the mass and stiffness modifications that were made to include operational and environmental variance (State#15-17). For the state settings with operational and environmental fluctuations, the mass and stiffness modifications are intended to ensure changes in the initial natural frequencies between a range of +/- 5 Hz.

## 4. Results and discussion

### 4.1 Proposed CNN model architecture

The dataset utilized in this study forms the bedrock upon which the investigation into structural damage detection using CNN rests. Each data sample within this dataset encapsulates the sensor readings, offering insight into various features that we denote as $X = \{X1, X2, X3, X4, X5, Time\}$. The crux of this dataset is the binary target variable y, which plays a pivotal role in distinguishing between two fundamental states of structures: damaged and undamaged. Each instance in this dataset equips with valuable real-world information about the health and status of the structure under scrutiny. Data preprocessing serves as the foundational layer for the structural damage detection framework presented in this study. This crucial phase comprises several intricate steps aimed at refining the raw sensor data, ensuring that it adheres to the prerequisites of subsequent model training. Two pivotal aspects of data preprocessing is dealt with: handling missing values and normalization.

Handling missing values is a prevalent challenge in the domain of data analysis. To address this issue, a straightforward yet effective imputation strategy is employed, involving the use of a simple imputer method designed to rectify missing values by replacing them with the mean value of each respective feature. The outcome is a dataset free from gaps or omissions in sensor readings, facilitating subsequent analytical steps. Mathematically, this can be expressed by Eq. (4):

$$X_{imputed} = Imputer(X) \quad \textbf{Eq. (4)}$$

Where, $X_{imputed}$ represents the dataset with missing values imputed.

Normalization plays a pivotal role in preprocessing, ensuring that feature values in the dataset adhere to a uniform and standardized numerical range. Its primary objective is to prevent specific features from dominating others during the model training phase. In this study, the well-established Min/Max scaling technique is adopted to normalize the sensor data. This involves rescaling feature values to fall within a common interval, typically [0, 1]. In practical terms, the original sensor readings are transformed by subtracting the minimum feature value and then dividing by the range, which is defined as the maximum feature value minus the minimum feature value as expressed by Eq. (5). This process ensures that each feature maintains a consistent numerical scale, making them suitable for the subsequent deep learning model's learning process.

$$X_{normalized} = \frac{X_{imputed} - \min(X_{imputed})}{\max(X_{imputed}) - \min(X_{imputed})} \quad \text{Eq. (5)}$$

Where, $X_{normalized}$ represents the normalized dataset. The preprocessed dataset, imbued with imputed values and normalized feature scales, forms the solid foundation upon which the CNN model for structural damage detection is erected. The significance of these preprocessing steps lies in their ability to set the stage for effective feature extraction, as well as ensuring the overall quality and consistency of the data used for training the model.

At the core of this research lies the application of CNNs as a potent tool for structural damage detection. CNNs, widely recognized for their effectiveness in addressing image-related tasks, are skilfully adapted in this study to analyse time-series sensor data. This section explores the specifics of the CNN model architecture utilized in this research, outlining the primary layers, their configurations, and the reasoning behind each design decision:

Input Layer: The CNN model commences with an input layer, functioning as the gateway for information entry into the neural network. In our particular architecture, the input layer has a dimension of (6, 1) to accommodate the six normalized sensor features. These features are supplied to the network as a one-dimensional (1D) array, facilitating smooth integration with the subsequent convolutional and fully connected layers.

Convolutional Layer: The convolutional layer is a crucial element responsible for extracting intricate patterns and feature representations from the input data. For the present work, a one-dimensional convolutional layer is utilized, featuring 32 filters with a kernel size of 3 as expressed in Eq. (6). The output of this layer is subjected to the ReLU (Rectified Linear Unit) activation function. This configuration selection enables the model to capture unique temporal patterns within the sensor data, enhancing the effectiveness of structural damage detection.

$$Output = Conv1D(32, 3, ReLU) \quad \text{Eq. (6)}$$

Max Pooling Layer: Following the convolutional layer, a max-pooling layer is integrated into the architecture. This layer plays a crucial role in down-sampling the feature maps generated by the convolutional layer. With a pool size of 2, the max-pooling layer condenses the feature representations, effectively reducing the dimensionality of the data as expressed in Eq. (7). This reduction not only expedites the subsequent computation but also ensures that the model remains robust in the face of varying time-series lengths:

$$Dimensionality = MaxPooling\ 1D(2) \quad \text{Eq. (7)}$$

Flatten Layer: The output from the max-pooling layer is then passed through a flatten layer. The purpose of this layer is to transform the two-dimensional data obtained from the previous layers into a one-dimensional vector format as expressed by Eq. (8). This 1D vector serves as the intermediary link between the convolutional layers and the densely connected layers, facilitating the seamless flow of information through the network:

$$Output = Flatten() \quad \text{Eq. (8)}$$

Dense layers: The architecture culminates with two fully connected (Dense) layers. These layers are essential in extracting high-level features from the data and performing the final classification of the structural health. The first dense layer comprises 16 neurons, each equipped with the ReLU activation function as expressed by Eq. (9). This configuration promotes the network's ability to capture complex relationships within the data:

$$First\ Dense\ layer = Dense(16, ReLU) \quad \text{Eq. (9)}$$

The final dense layer is a single neuron that employs the sigmoid activation function. This layer is pivotal for producing the ultimate binary output, signifying whether the structural system under observation is in a damaged state or remains undamaged as expressed by Eq. (10):

$$Final\ Dense\ Layer = Dense(1, Sigmoid) \quad \textbf{Eq. (10)}$$

Training Process: The training process is a pivotal aspect of the methodology, enabling the model to acquire knowledge from the dataset and perform structural damage detection. This section delves into the specifics of training the Convolutional Neural Network (CNN) model, outlining the loss function, optimizer, and other crucial training parameters. The core of the training process centers on the loss function, a crucial element that measures the difference between the model's predictions and the actual ground truth labels. For binary classification task, the binary cross-entropy loss function, commonly referred to as log loss, is utilized. This loss function is especially suitable for binary classification challenges and serves as a key factor in steering the model's learning process. Its mathematical representation is given by Eq. (11):

$$Loss\ Function: Binary\ Cross - Entropy = \\ -(y.\log(\hat{y}) + (1-y).\log(1-\hat{y})) \quad \textbf{Eq. (11)}$$

Where $y$ represents the true label, and $\hat{y}$ denotes the predicted probability of the machinery or structural system being in a damaged state. The binary cross-entropy loss function provides a measure of how closely the model's predictions align with the actual outcomes, thereby steering the iterative training process.

To optimize the model during training, the Adam optimizer is employed. Adam is an adaptive learning rate optimization algorithm that efficiently updates the model's weights during back propagation. Its dynamic learning rate adjustments facilitate rapid and effective convergence to a configuration that minimizes the loss function. The selection of Adam as the optimizer is based on its established efficacy in training deep neural networks across various applications. Training parameters, including the number of epochs and batch size, hold a significant influence on the training process. The model undergoes training over a predetermined number of epochs, indicating the total passes through the dataset. In this study, ten epochs are performed, affording numerous chances for the model to refine its weight parameters and improve its predictive accuracy. Additionally, within each epoch, the dataset is partitioned into smaller batches, and the model's weights are updated incrementally after handling each batch. A batch size of 16 is employed in this study, striking a balance between computational efficiency and the effectiveness of weight updates as expressed by Eq. (12).

$$Model.fit(X_{train}, y_{train}, epochs = 10, batch_{size} = 16) \\ \textbf{Eq. (12)}$$

The performance of the CNN model, designed for structural damage detection, is rigorously evaluated using a well-established evaluation metric: accuracy. Accuracy serves as a quintessential measure of the model's effectiveness in correctly classifying instances within the dataset. It is defined as the ratio of the number of correct predictions to the total number of predictions, expressed as Eq. (13):

$$Accuracy = \frac{Number\ of\ Correct\ Predictions}{Total\ Number\ of\ Predictions} \quad \textbf{Eq. (13)}$$

The accuracy metric offers a comprehensive assessment of the model's ability to discern between damaged and undamaged states in structural systems. A higher accuracy score underscores the model's proficiency in making accurate predictions, validating its utility for real-world applications in structural health monitoring.

Once the CNN model has been meticulously trained and attains a satisfactory level of accuracy, it is poised for real-world applications, such as the detection of structural damage in newly acquired sensor data. This section delineates the step-by-step procedure for making predictions on new data, providing a glimpse into the practical deployment of the trained model.

To make predictions on new data, the following steps are followed:

The predictive capabilities of the model come into play when provided with new sensor data. In the context of our study, this new sensor data comprises readings from various sensors, including Sensor 1, Sensor 2, Sensor 3, Sensor 4, Sensor 5, and Time. This set of sensor readings represents the real-time data collected from the structural system under observation, offering insights into its current state. After handling missing values and normalizing the new data, it is crucial to reshape the data to match the input shape expected by the CNN. In the present model architecture, the CNN's input layer is configured to accept a one-dimensional (1D) array of sensor features. The reshaping process involves transforming the normalized data into the desired format, ensuring that the data's dimensions align with the model's expectations (1,6,1). The culmination of the predictive process involves utilizing the fully trained model to predict the probability of the structural system being in a damaged state as expressed by Eq. (14). With the reshaped and normalized new sensor data as input, the model processes this information through its layers and computes the probability score that corresponds to structural damage. This probability score is obtained from the final layer, which employs the sigmoid activation function to yield a binary output.

$Prediction = Model.predict(X_{new\_reshaped})$ **Eq. (14)**

In practical terms, the model's prediction output signifies the likelihood of the structural system being in a damaged state. The continuous nature of this probability score can be interpreted and thresholded to determine a binary classification, thereby offering a real-time assessment of the structural health. These detailed steps provide a comprehensive understanding to the methodology used in this study for the structural damage detection using CNN.

### 4.2 Output obtained

The graphical output as obtained in Fig. 3 portrays a critical aspect of the model's training process. Here, the x-axis serves as a timeline, tracking the number of training epochs undergone, while the y-axis accommodates two distinct curves: one for training accuracy and the other for validation accuracy. These accuracy metrics gauge the model's performance during different phases of its training. The evaluation criterion employed for both metrics is accuracy, a measure of the model's capability to correctly classify instances within the data. The training accuracy curve traces the evolution of the model's performance with respect to the training data. In the early epochs, it is customary to observe a progressive increase in training accuracy as the model acquaints itself with the training dataset. However, it is imperative to recognize that high training accuracy, in isolation, does not guarantee the model's ability to generalize effectively to new, unseen data. Complementing the training accuracy curve, the validation accuracy curve scrutinizes the model's proficiency when faced with previously unencountered data. An upturn in validation accuracy during the initial stages of training signifies the model's acquisition of valuable patterns. Nevertheless, it is paramount to remain vigilant for any divergence that might manifest between the training and validation accuracy curves, as such disparities can signal a phenomenon known as overfitting, wherein the model becomes excessively attuned to the training data, potentially compromising its performance on new data. Together, the combined accuracy versus epoch graph offers a holistic insight into the model's training dynamics, serving as a valuable tool for assessing its adaptability and predictive capabilities.

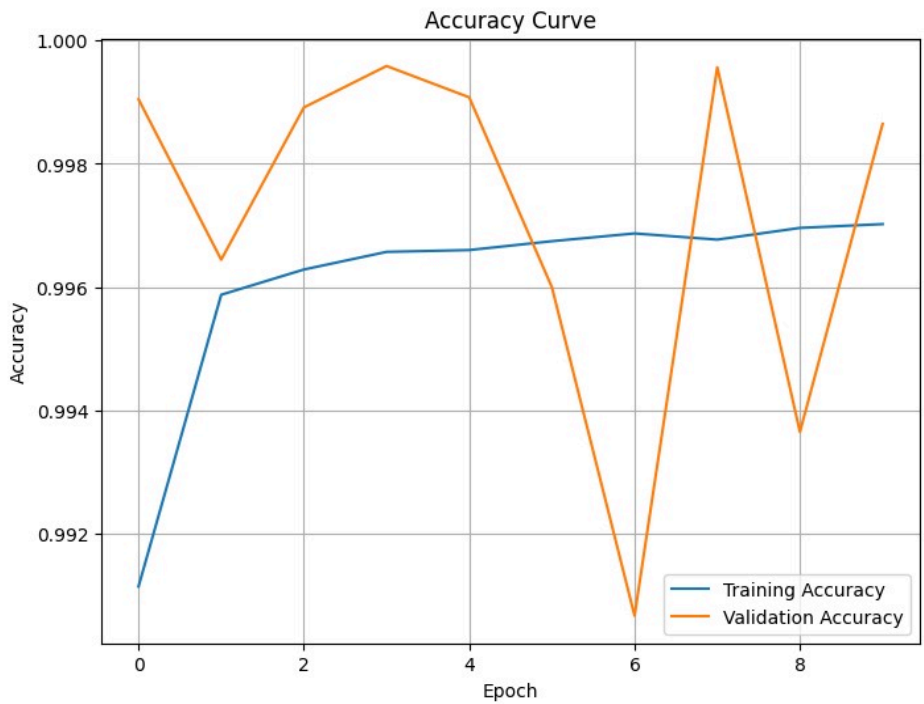

**Fig. 3:** Accuracy-Epoch curve.

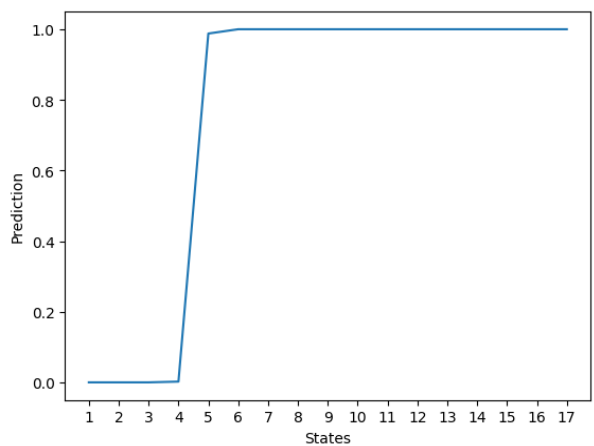
(a)

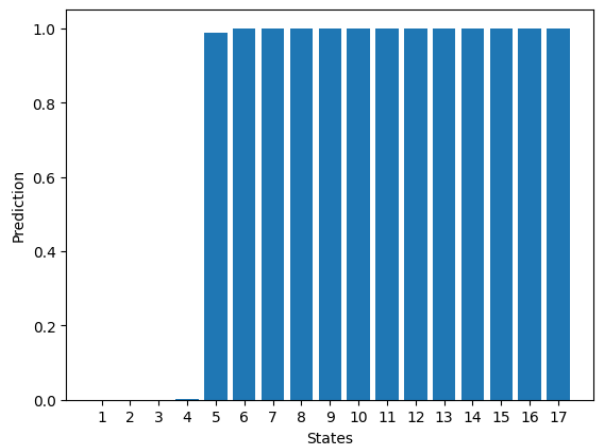
(b)

**Fig. 4:** State curve

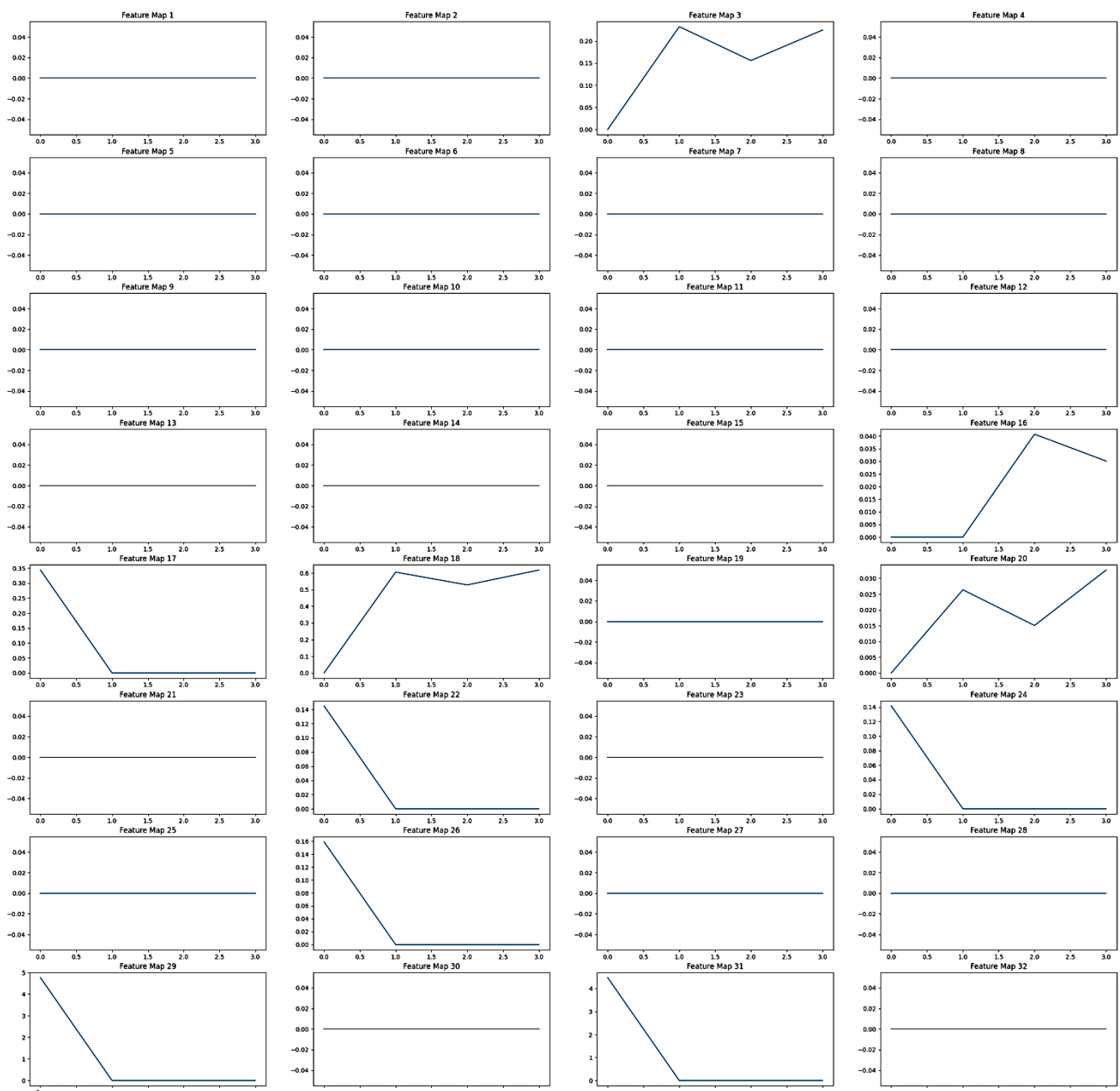

**Fig. 5:** Feature maps

The mean predictions graph depicted in Fig. 4 offers an insightful perspective on the predictive performance of the CNN model across all 17 states. In the context of each state, the CNN model generates a prediction score denoting the probability of structural damage. These individual predictions are then subjected to averaging to yield the mean prediction for each respective state. The x-axis of the graph corresponds to the 17 states, while the y-axis represents the mean prediction scores. This visual representation serves to elucidate the model's proficiency in discriminating between states characterized by structural damage and those in an undamaged condition. Notably, higher mean prediction scores correspond to a heightened likelihood of damage. The mean predictions graph serves as compelling evidence of the CNN model's capacity to effectively distinguish between states featuring structural damage and those devoid of such issues as shown in Fig 4(b). It is particularly evident in the case of States 1 to 3, deliberately subjected to damage, as they manifest significantly elevated mean prediction scores in contrast to their undamaged counterparts. This graphical representation underscores the model's prowess in classifying machine states based on the information gleaned from sensor data.

Fig. 5 shows the feature maps. Each subplot in the figure corresponds to a different feature map, identified as "Feature Map 1," "Feature Map 2," and so on. The X-axis of each subplot represents the position along the input sequence (spatial dimension), while the Y-axis represents the strength or activation of the feature detected at each position. Higher values on the Y-axis indicate a stronger presence of the feature, while lower values indicate a weaker presence. By examining these feature map activations, one can gain insights into what local patterns or features the Conv1D layer has learned from the input data. Features with higher activations are more important for the layer's decision-making process. These visualizations are useful for debugging and understanding the inner workings of the Conv1D model and can help identify the types of patterns it's sensitive to.

## 5. Conclusion

Present research demonstrates the effectiveness of CNNs in the domain of structural damage detection using time-series sensor data. The proposed CNN model architecture, comprised of key layers such as Convolutional, Max Pooling, Flatten, and Dense layers, exhibits the capability to capture intricate temporal patterns in sensor data. Through training with binary cross-entropy loss and optimization with the Adam optimizer, the model generalizes well to unseen data, as evident in training and validation accuracy curves. The mean predictions graph highlights the model's proficiency in distinguishing between states with and without structural damage, offering valuable insights for real-world structural health monitoring applications. This research contributes technically to the field and opens avenues for future work in more complex architectures and real-time monitoring systems for structural health assessment. This study showcases the potential of CNNs in effectively addressing structural damage detection, a critical concern in the realm of structural engineering. The methodology, findings, and technical contributions presented here set the stage for further advancements in structural health monitoring, with the aim of enhancing the safety and reliability of various structural systems.